\documentclass[11pt]{article}
\usepackage[]{acl}
\usepackage{balance}
\usepackage{times}
\usepackage{latexsym}

\usepackage[T1]{fontenc}

\usepackage[utf8]{inputenc}

\usepackage{microtype}

\usepackage{xspace}
\usepackage{paralist}
\usepackage{enumitem}
\usepackage{graphicx}
\usepackage{multirow}
\usepackage{booktabs}
\usepackage{subfig}
\usepackage{xcolor}
\usepackage{hyperref}

\definecolor{E66826}{HTML}{E66826}
\definecolor{345AB5}{HTML}{345AB5}

\newcommand*{\eqref}[1]{~(\ref{#1})}

\newcommand{\method}{NEP\xspace}
\newcommand{\authors}{Qiang Sheng, Juan Cao\thanks{$^*$Corresponding author.}, Xueyao Zhang, Rundong Li, Danding Wang, Yongchun Zhu}
\newcommand{\emailsI}{\href{mailto:shengqiang18z@ict.ac.cn}{shengqiang18z},\href{mailto:caojuan@ict.ac.cn}{caojuan},\href{mailto:zhangxueyao19s@ict.ac.cn}{zhangxueyao19s}}
\newcommand{\emailsII}{\href{mailto:lirundong20s@ict.ac.cn}{lirundong20s},\href{mailto:wangdanding@ict.ac.cn}{wangdanding},\href{mailto:zhuyongchun18s@ict.ac.cn}{zhuyongchun18s}}

\title{\textit{Zoom Out and Observe}: \\ News Environment Perception for Fake News Detection}
\author{\authors\\
	Key Lab of Intelligent Information Processing of Chinese Academy of Sciences, \\
	Institute of Computing Technology, Chinese Academy of Sciences\\
	University of Chinese Academy of Sciences\\
	\texttt{\{\emailsI\}@ict.ac.cn}\\
	\texttt{\{\emailsII\}@ict.ac.cn}
}

\begin{document}
\maketitle

\begin{abstract}

Fake news detection is crucial for preventing the dissemination of misinformation on social media.
To differentiate fake news from real ones, existing methods observe the language patterns of the news post and ``zoom in'' to verify its content with knowledge sources or check its readers' replies.
However, these methods neglect the information in the external \emph{news environment} where a fake news post is created and disseminated.
The news environment represents recent mainstream media opinion and public attention, which is an important inspiration of fake news fabrication because fake news is often designed to ride the wave of popular events and catch public attention with unexpected novel content for greater exposure and spread.
To capture the environmental signals of news posts, we ``zoom out'' to observe the news environment and propose the News Environment Perception Framework (\method).
For each post, we construct its macro and micro news environment from recent mainstream news. Then we design a popularity-oriented and a novelty-oriented module to perceive useful signals and further assist final prediction.
Experiments on our newly built datasets show that the NEP can efficiently improve the performance of basic fake news detectors.\footnote{\url{https://github.com/ICTMCG/News-Environment-Perception/}}
\end{abstract}

\section{Introduction}
The wide spread of fake news on online social media has influenced public trust~\cite{Knight} and poses real-world threats on politics~\cite{pizzagate}, finance~\cite{obama}, public health~\cite{infodemic}, etc. Under such severe circumstances, automatically detecting fake news has been an important countermeasure in practice.

\begin{figure}[t]
\setlength{\belowcaptionskip}{-0.5cm}
	\centering
	\includegraphics[width=\linewidth]{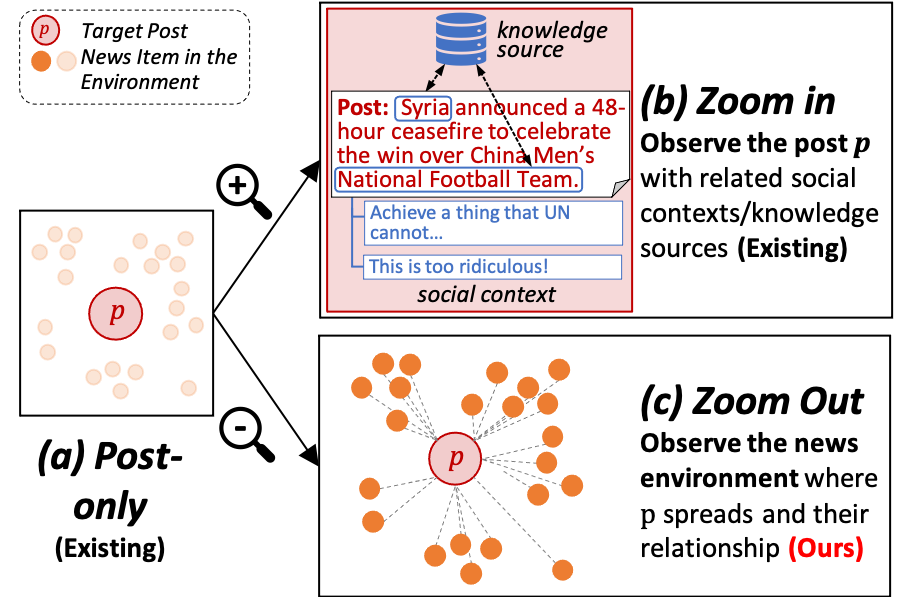}
	\caption{Existing methods for fake news detection rely on (a) the post content itself and (b) related post-level signals like social context and knowledge. Unlike (a) and (b), our method captures (c) signals from \emph{news environments}.}
	\label{fig:motivating}
\end{figure}

Besides directly observing the post's content patterns~\cite{fiction, eann} (\figurename~\ref{fig:motivating}(a)), most existing methods for fake news detection ``zoom in'' for finding richer post-level signal by checking user replies to the post~\cite{defend, dual-emotion} and verifying the claim with knowledge sources~\cite{declare, kmgcn} (\figurename~\ref{fig:motivating}(b)).	
However, these methods neglect a different line of ``zooming out'' to observe the external \emph{news environment} where a fake news post is created and disseminated.
Our starting point is that a news environment, which represents recent mainstream media opinion and public attention, is an important inspiration of the fabrication of contemporary fake news.
Since any gains of fake news achieve only if it widely exposes and virally spreads, a fake news creator would carefully design how to improve the post's visibility and attract audiences' attention in the context (environment) of recently published news.
Such intentional design connects fake news with its news environment and conversely, we might find useful signals from the news environment to better characterize and detect fake news.

\begin{figure}[t]
\setlength{\belowcaptionskip}{-0.5cm}
	\centering
	\includegraphics[width=\linewidth]{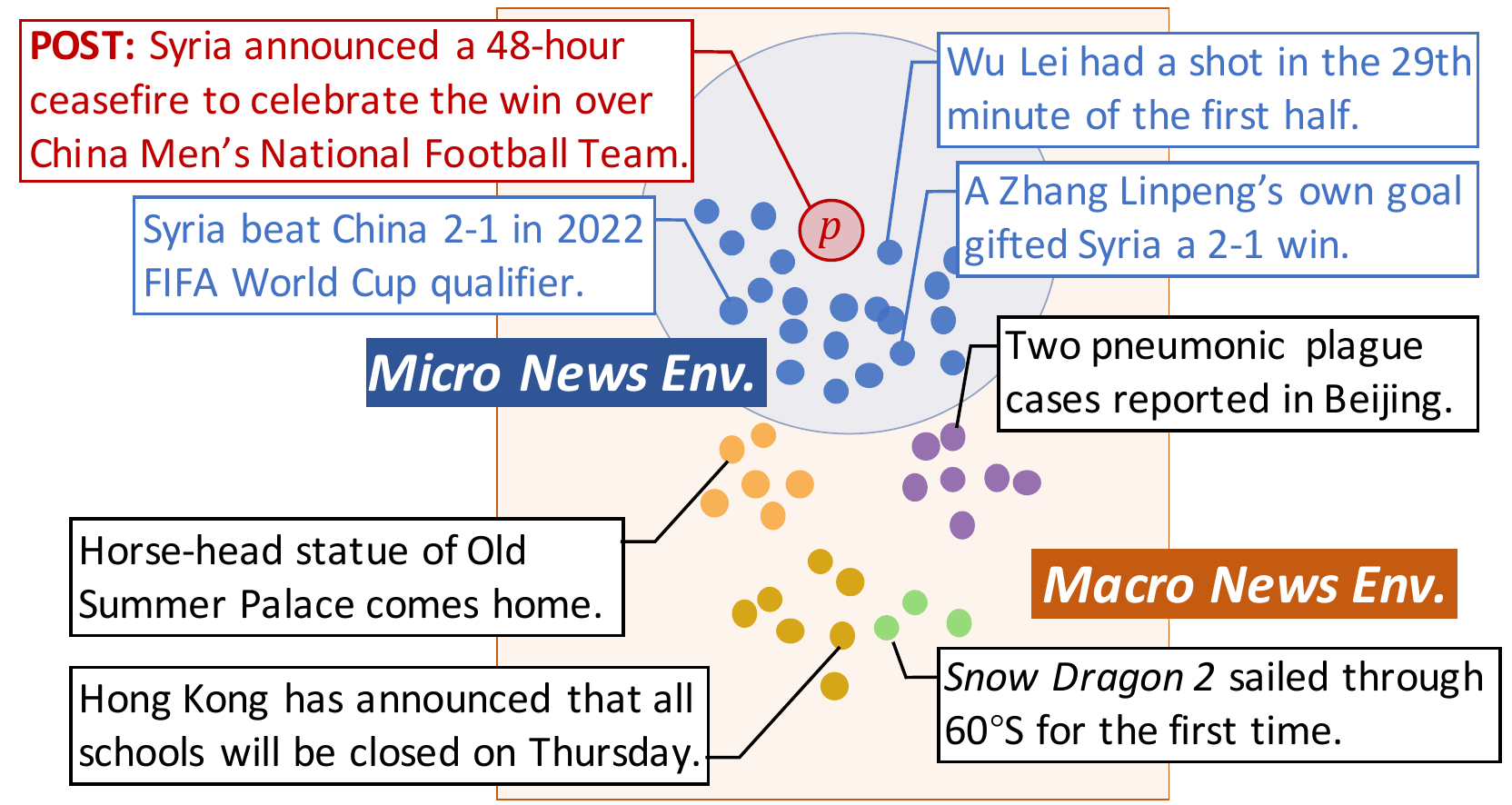}
	\caption{A fake news post $p$ and its news environment containing recent news items in three days (2019/11/12 to 2019/11/14). Only the items in events that are reported multiple times (differentiated by dot colors) are displayed for brevity. We can see that $p$ falls in a \emph{popular} event on a Syria-China World Cup qualifier compared with other events and focuses on a \emph{novel} aspect (unusual celebration in Syria).}
	\label{fig:eg}
\end{figure}

\figurename~\ref{fig:eg} shows an example, where we name the whole set of recent news items the \emph{macro} news environment and the event-similar subset as the \emph{micro} news environment.
For the fake news post $p$ on Syria's ceasefire thanks to a win over China in a football match, we observe two important signals from its news environments:
 
 \textbf{1) Popularity.} In the \emph{macro} news environment that contains all recent news items, $p$ is related to a relatively popular event (Syria-China football match) among the five events in different domains. This would bring $p$ greater exposure and further greater impact.
 
 \textbf{2) Novelty.} In the \emph{micro} news environment, the items mostly focus on the game itself (e.g., ``Wu Lei had a shot''), while $p$ provides novel side information about Syria's unusual celebration. This would help catch audiences' attention and boost the spread of $p$~\cite{science18}.
 
 Unfortunately, these potentially useful signals could be hardly considered by post-only and ``zoom-in'' methods, as they focus on digging in the direction towards inherent properties of a single post (e.g., styles, emotions and factual correctness), rather than observing the surrounding environments of the post.

\begin{figure*}[htbp]
\setlength{\belowcaptionskip}{-0.2cm}
	\centering
	\includegraphics[width=\textwidth]{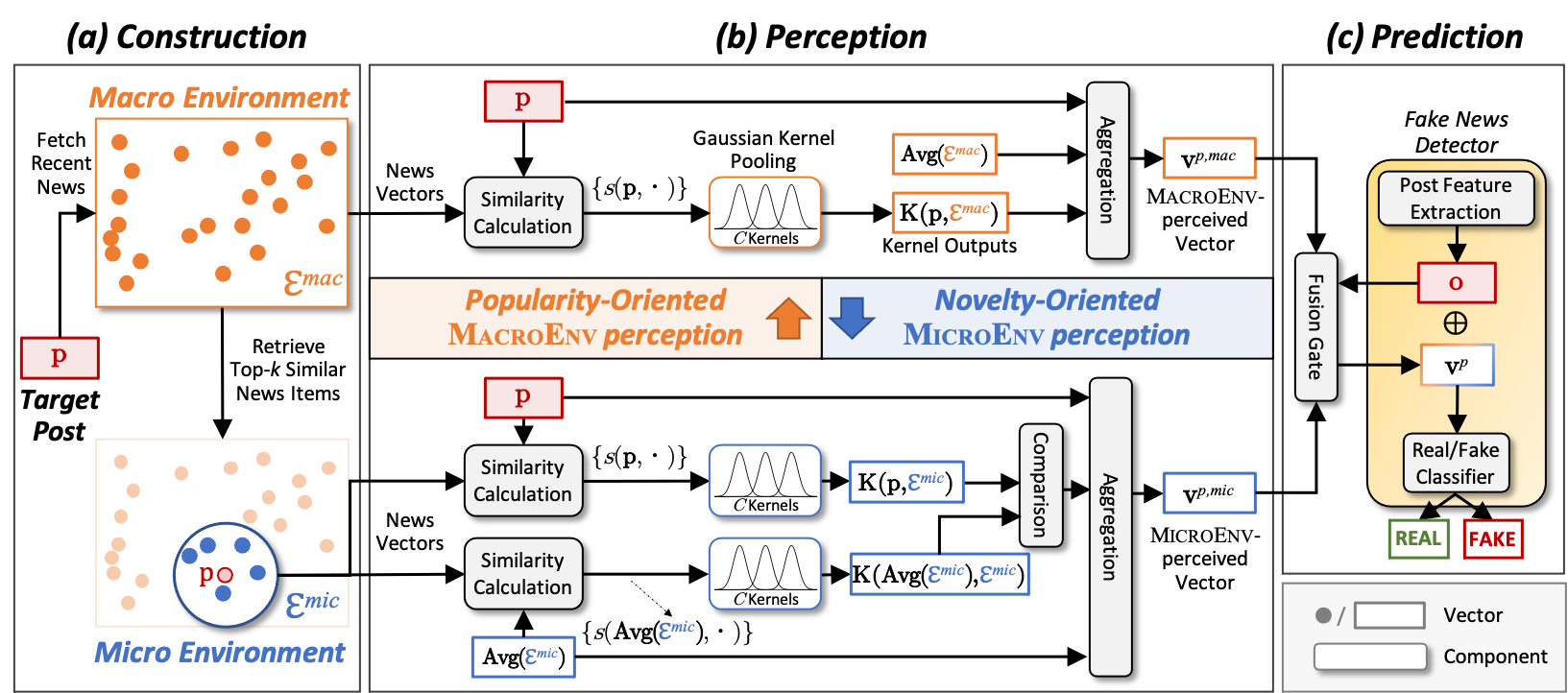}
	\caption{Architecture of the News Environment Perception Framework (NEP).
	\textbf{(a) Construction:} The macro and micro news environments (\textsc{MacroEnv} and \textsc{MicroEnv}) of the target post $p$ (whose representation vector at the construction and perception stages is $\mathbf{p}$) are constructed using recent mainstream news items.
	\textbf{(b) Perception:} We perceive $p$'s popularity in the \textsc{MacroEnv} and novelty in the \textsc{MicroEnv} based on the outputs of Gaussian Kernel Pooling ($\mathbf{K}$) which counts on similarities ($\{s(\cdot,\cdot)\}$) in a soft manner. This results in two environment-perceived vectors, $\mathbf{v}^{p,mac}$ and $\mathbf{v}^{p,mic}$.
	\textbf{(c) Prediction:} Environment-perceived Vectors are fused with a gate guided by the extracted post feature $\mathbf{o}$ (does not necessarily equal $\mathbf{p}$) from the fake news detector such as EANN~\cite{eann}, BERT~\cite{bert}, and others. Fused $\mathbf{v}^{p}$ and $\mathbf{o}$ are fed into the final classifier (typically, an MLP) for prediction of $p$ being fake or real.
	}
	\label{fig:arch}
\end{figure*}

To enable fake news detection systems to exploit information from news environments, we propose the News Environment Perception Framework (NEP). As presented in \figurename~\ref{fig:arch}, for the post $p$, we construct two news environments, \textsc{MacroEnv} and \textsc{MicroEnv}, using recent mainstream news data to facilitate the perception from different views.
We then design a popularity-oriented and a novelty-oriented perception module to depict the relationship between $p$ and these recent news items.
The environment-perceived vectors are fused into an existing fake news detector for prediction.

Our contributions are as follows:
\begin{compactitem}
	\item \textbf{Problem}: To the best of our knowledge, we are the first to incorporate news environment perception in fake news detection.
	\item \textbf{Method}: We propose the \method framework which exploits the perceived signals from the macro and micro news environments of the given post for fake news detection.
	\item \textbf{Data \& Experiments}: We construct the first dataset which includes contemporary mainstream news data for fake news detection. Experiments on offline and online data show the effectiveness of NEP.
\end{compactitem}



%

\section{Related Work}
Fake news detection is mostly formulated as a binary classification task where models are expected to accurately judge the given post as real or fake. Existing works focus on discovering distinctive features \emph{in} the post from various aspects as Figure~\ref{fig:eg} shows, which we roughly group them as:

\textbf{Post-only methods} aim at finding shared patterns in appearances across fake news posts (\figurename~\ref{fig:motivating}(a)). Text-based studies focus on better constructing features based on sentiment~\cite{ajao}, writing style~\cite{style-aaai}, language use~\cite{fiction}, discourse~\cite{hdsf}, etc. Other works rely on deep neural models to encode contents and handle certain scenarios, such as visual-based~\cite{qipeng-icdm, qipeng-chapter}, multi-modal~\cite{eann, qipeng-mm} and multi-domain~\cite{mdfend} detection. Our NEP provides additional news environmental information and can coordinate with post-only methods (will show in Section~\ref{sec:exp}).

\textbf{``Zoom-in'' methods} introduce related sources to understand the post delicately.
One line is to use \emph{social contexts} (bottom of \figurename~\ref{fig:motivating}(b)). Some directly analyze the network information to find patterns shaped by user relationship and information diffusion~\cite{shu-beyond,network-zhou,fang,Propagation2Vec}, and others leverage collective wisdom reflected by user responses~\cite{majing-tree,allinone,defend,dual-emotion}. For example, a refuting reply saying \textit{``FYI, this is false''} would be an important reference to make a prediction.
Another line refers to \emph{knowledge sources} (top of \figurename~\ref{fig:motivating}(b)) and aims at verifying the post with retrieved evidence for detection. The knowledge sources can be webpages~\cite{declare, majing-claim, mac, cicd, pref-fend}, knowledge graphs~\cite{deterrent}, online encyclopedias~\cite{fever, feverous}, fact-checking article bases~\cite{multifc, shaar}, etc.
Our NEP starts from a different view, for it ``zooms out'' to observe the news environment where the post spreads. Note that our method is \emph{not equivalent} to a knowledge-based method that uses news environments as evidence bases, as it does not pick evidential news items to prove or disprove the given post, but aims at reading the news ``atmosphere'' when the post is published. 
In that sense, ``zoom-in'' and ``zoom-out'' methods can actually be integrated for comprehensively detecting fake news (will also show in Section~\ref{sec:exp}).

\section{Proposed Method}
\figurename~\ref{fig:arch} overviews our proposed framework NEP, whose goal is to empower fake news detectors with the effective perception of news environments. Given a post $p$, we first construct its macro and micro environment (\textsc{MacroEnv} and \textsc{MicroEnv}) using recent news data.
Then we model the post-environment relationships to generate environment-perceived vectors $\mathbf{v}^{p,mac}$ and $\mathbf{v}^{p,mic}$.
Finally, the two vectors are fused with post representation $\mathbf{o}$ derived from the fake news detector to predict if $p$ is real or fake.
\subsection{News Environment Construction}
The environment is the objects, circumstances, or conditions by which one is surrounded~\cite{environment}.
Accordingly, a news environment should contain news reports which can reflect the present distribution of mainstream focuses and audiences' attention.
To this end, we collect news items published by \textit{mainstream media outlets} as basic environmental elements, in that their news reports generally face a large, common audience.

Let $\mathcal{E}$ be the set of all collected news items published earlier than $p$.
We construct a macro environment (\textsc{MacroEnv}) and a micro environment (\textsc{MicroEnv}), which are defined as follows:

\begin{compactitem}
	\item \textsc{MacroEnv} is the set of news items in $\mathcal{E}$ released within $T$ days before $p$ is published:
	\begin{equation}
		\mathcal{E}^{mac}=\{e: e\in\mathcal{E}, 0 < t_p - t_e \leq T\},
	\end{equation}
	where $t_p$ and $t_e$ respectively denote the publication date of $p$ and the news item $e$.
	\item \textsc{MicroEnv} is the set of news items in $\mathcal{E}^{mac}$ that are relevant to $p$. Here, we query $\mathcal{E}^{mac}$ using $p$ and obtain the top $k$ as the set:
	\begin{equation}
		 \mathcal{E}^{mic}=\{e: e\in \mathrm{Topk}(p, \mathcal{E}^{mac})\},
	\end{equation}
	where $k=\lceil r|\mathcal{E}^{mac}|\rceil$ and $r \in (0,1)$ determines the proportion.
\end{compactitem}

Intuitively, the time-constrained environment \textsc{MacroEnv} provides a macro perspective of what the mass audience read and focus on recently, while the further relevance-constrained one \textsc{MicroEnv} describes the distribution of items about similar events. We use a pretrained language model $\mathcal{M}$ (e.g., BERT~\cite{bert}) to obtain the post/news representation. For $p$ or each item in the macro/micro environment $e$, the initial representation is the output of $\mathcal{M}$ for the $\mathtt{[CLS]}$ token:
\begin{equation}
	\mathbf{p}=\mathcal{M}(p),\ \ \mathbf{e}=\mathcal{M}(e).
\end{equation} 


\subsection{News Environment Perception}
\label{sec:perception}
The perception of news environments of $p$ is to capture useful signals from existing mainstream news items. The signals are expected to discover unique post-environment interactive patterns of fake news. Starting from the motivation of fake news creators to widely diffuse fabricated information to the whole online news ecosystem, we guide the model to perceive from two important diffusion-related perspectives, i.e., popularity and novelty, in the \textsc{MacroEnv} and the \textsc{MicroEnv}. 

\noindent\textbf{Popularity-Oriented \textsc{MacroEnv} Perception.}
A fabricated post would be more likely to go viral and thus gain more influence when it is related to trending news. Thus, a fake news creator might consider how to chase clouts of hot events during writing a fake news post.
Here we consider how popular the main event of $p$ is in the \textsc{MacroEnv}.
We transform the perception of popularity into the similarity estimation between $p$ and individual news items. That is, if many items in the \textsc{MacroEnv} are similar to $p$, then $p$ might be also popular in such an environment.
Following~\cite{sentence-bert}, we first calculate cosine similarity between $\mathbf{p}$ and each news item (say, $i$) in $\mathcal{E}^{mac}$:
\begin{equation}
	s(\mathbf{p}, \mathbf{e}_i) = \frac{\mathbf{p}\cdot\mathbf{e}_i}{\Vert\mathbf{p}\Vert\Vert\mathbf{e}_i\Vert}.
\end{equation}

The similarity list $\{cos(\mathbf{p}, \mathbf{e}_i)\}_{i=1}^{|\mathcal{E}^{mac}|}$ of variable length $|\mathcal{E}^{mac}|$ does not work well with networks mostly taking fixed-dimensional vectors as inputs. Thus, the list requires a further transformation, where we expect the transformed environment-perceived vector to reflect how similar $p$ is to the environment without much information loss. Following~\cite{kernelpooling,kgat}, we here choose to calculate a soft \emph{counting} on the list to obtain a distribution that mimics a hard bin plot. Specifically, we employ a Gaussian Kernel Pooling proposed in~\cite{kernelpooling} across the range of cosine similarity to get soft counting values. Assuming that we use $C$ kernels $\{\mathbf{K}_i\}_{i=1}^{C}$, the output of $k$-th kernel is:

\begin{equation}
    \mathbf{K}_{k}^{i}= \exp \left({-\frac{(s(\mathbf{p},\mathbf{e}_i) - \mu_k)^2}{2\sigma_k^2}}\right),
    \label{eq:kernel_i}
\end{equation}
\begin{equation}
	\mathbf{K}_{k}(\mathbf{p}, \mathcal{E}^{mac}) = \sum_{i=1}^{|\mathcal{E}^{mac}|}\mathbf{K}_{k}^{i},
	\label{kernel}
\end{equation}
where $\mu_k$ and $\sigma_k$ is the mean and width of the $k$-th kernel.  In Eq.~\eqref{eq:kernel_i}, if the similarity between $\mathbf{p}$ and $\mathbf{e}$ is close to $\mu_k$, the exponential term will be close to 1; otherwise to 0. We then sum the exponential terms with Eq.~\eqref{kernel}. This explains why a kernel is like a soft counting bin of similarities. We here scatter the means $\{\mu_k\}_{k=1}^{C}$ of the $C$ kernels in $[-1,1]$ to completely and evenly cover the range of cosine similarity. The widths are controlled by $\{\sigma_k\}_{k=1}^{C}$. Appendix~\ref{app:kernel} provides the details. A $C$-dim similarity feature in the \textsc{MacroEnv} is obtained by concatenating all kernels' outputs and normalizing with the summation of the outputs:
\begin{equation}
	\mathbf{K}(\mathbf{p}, \mathcal{E}^{mac}) \!=\! \mathrm{Norm}\left( \bigoplus_{k=1}^{C} \mathbf{K}_{k}(\mathbf{p}, \mathcal{E}^{mac})\right),
\label{kernels}
\end{equation}
where $\bigoplus$ is the concatenation operator and $\mathrm{Norm}(\cdot)$ denotes the normalization.

By calculating $\mathbf{K}(\mathbf{p}, \mathcal{E}^{mac})$, we obtain a soft distribution of similarities between $p$ and the \textsc{MacroEnv} as the perception of popularity. To enrich the perceived information, we generate the \textsc{MacroEnv}-perceived vector for $p$ by fusing the similarity and semantic information. Specifically, we aggregate the post vector, the center vector of the \textsc{MacroEnv} $\mathbf{m}(\mathcal{E}^{mac})$ (by averaging all vectors), and the similarity feature using an MLP:
\begin{equation}
	\mathbf{v}^{p,mac}\!=\! \mathrm{MLP}(\mathbf{p} \oplus \mathbf{m}(\mathcal{E}^{mac}) \oplus \mathbf{K}(\mathbf{p}, \mathcal{E}^{mac})\!).
\end{equation}

\noindent\textbf{Novelty-Oriented \textsc{MicroEnv} Perception.}
Different from \textsc{MacroEnv}, \textsc{MicroEnv} contains mainstream news items close to $\mathbf{p}$, which indicates that they are likely to share similar events. However, even in a popular event, a post may still be not attended if it is \emph{too} similar to others. \citet{science18} found that false news was more novel than true news on Twitter with the reference to the tweets that the users were exposed to (could be regarded as a user-level news environment). This might explain why fake news spread ``better''. We thus consider how novel $p$ is in the event-similar \textsc{MicroEnv}.\footnote{We perceive the novelty in the \textsc{MicroEnv} rather than the \textsc{MacroEnv} to mitigate the effects of event shift.}

If the content of a post is novel, it is expected to be an outlier in such an event.
Here, we use the center vector $\mathbf{m}(\mathcal{E}^{mic})$ of \textsc{MicroEnv} as a reference. Specifically, we again use Eqs. \eqref{eq:kernel_i} to~\eqref{kernels}, but here, calculate \emph{two} similarity features $\mathbf{K}(\mathbf{p}, \mathcal{E}^{mic})$ and $\mathbf{K}(\mathbf{m}(\mathcal{E}^{mic}), \mathcal{E}^{mic})$. The latter serves as a reference for the former and facilitates the model ``calibrate'' its perception.
The generation of the \textsc{MicroEnv}-perceived vector for $p$ is as follows:
\begin{equation}
	\mathbf{u}^{sem}=\mathrm{MLP}(\mathbf{p} \oplus \mathbf{m}(\mathcal{E}^{mic})),
\end{equation}
\begin{equation}
	\mathbf{u}^{sim}\!=\!\mathrm{MLP}(\mathrm{g}(\mathbf{K}(\mathbf{p}, \mathcal{E}^{mic}), \!\mathbf{K}(\mathbf{m}(\mathcal{E}^{mic}), \mathcal{E}^{mic}))),
\end{equation}
\begin{equation}
	\mathbf{v}^{p,mic}= \mathrm{MLP}(\mathbf{u}^{sem} \oplus \mathbf{u}^{sim}),
\end{equation}
where the comparison function $\mathrm{g}(\mathbf{x},\mathbf{y})=(\mathbf{x} \odot \mathbf{y}) \oplus (\mathbf{x}-\mathbf{y})$ and $\odot$ is the Hadamard product operator. $\mathbf{u}^{sem}$ and $\mathbf{u}^{sim}$ respectively aggregate the semantic and similarity information. The $\mathrm{MLP}$s are individually parameterized. We omit their index numbers in the above equations for brevity.




\subsection{Prediction under Perceived Environments}
As our environment perception does not necessarily depend on a certain detection model, we expect our NEP to have a good compatibility with various fake news detectors. In our NEP, we achieve this by gate fusion.
Take a post-only detector as an example. We apply the gate mechanism for adaptively fusing $\mathbf{v}^{p,mac}$ and $\mathbf{v}^{p,mic}$ according to $\mathbf{o}$:
\begin{equation}
	\mathbf{v}^{p} = \mathbf{g}\odot\mathbf{v}^{p,mac}+(\mathbf{1}-\mathbf{g})\odot\mathbf{v}^{p,mic},
\end{equation}
where the gating vector $\mathbf{g}=\mathrm{sigmoid}(\mathrm{Linear}(\mathbf{o}\oplus\mathbf{v}^{p,mac}))$, $\mathrm{sigmoid}$ is to constrain the value of each element in $[0,1]$, and $\mathbf{o}$ denotes the last-layer feature from a post-only detector.\footnote{Empirically, we take the output of one of the last few dense layers whose dimensionality is moderate.} $\mathbf{o}$ and $\mathbf{v}^{p}$ are further fed into an MLP and a $\mathrm{softmax}$ layer for final prediction:
\begin{equation}
	\mathbf{\hat{y}}=\mathrm{softmax}(\mathrm{MLP}(\mathbf{o}\oplus\mathbf{v}^{p})).
	\label{pred}
\end{equation}
When working with more complex detectors that rely on other sources besides the post, we can simply concatenate those feature vectors in Eq.\eqref{pred}. For example, we can concatenate $\mathbf{v}^{p}$ with the post-article joint representation if the fake news detector is knowledge-based.
During training, we minimize the cross-entropy loss.

\section{Experiment}
\label{sec:exp}
We conduct experiments to answer the following evaluation questions:
\begin{compactitem}
 \item \textbf{EQ1:} 
 Can NEP improve the performance of fake news detection?
 \item \textbf{EQ2:} How effective does the NEP model the macro and micro news environments?
 \item \textbf{EQ3:} In what scenarios do news environments help with fake news detection?
\end{compactitem}

\subsection{Datasets}

We integrated existing datasets in Chinese and English and then collected news items released in the corresponding time periods. The reasons why we do not use a single, existing dataset include 1) no existing dataset provides the contemporary news items of verified news posts to serve as the elements in news environments; 2) most datasets were collected in a short time period and some suffer from a high class imbalance across years.\footnote{For example, Weibo-20~\cite{dual-emotion} is roughly balanced as a whole but has a ratio of 5.2:1 between real and fake news samples in 2018.} The statistics are shown in \tablename~\ref{data} and the details are as follows:


\begin{table}[t]
\setlength{\abovecaptionskip}{0.3cm}
\small
\centering
\setlength{\tabcolsep}{2pt}
\caption{Statistics of the datasets.}
\begin{tabular}{@{}crllllll@{}}
\toprule
\multicolumn{2}{c}{\multirow{2}{*}[-0.3em]{\textbf{Dataset}}} & \multicolumn{3}{c}{\textbf{Chinese}} & \multicolumn{3}{c}{\textbf{English}} \\
\cmidrule(){3-8} 
\multicolumn{2}{c}{} & \multicolumn{1}{c}{Train} & \multicolumn{1}{c}{Val} & \multicolumn{1}{c}{Test} & \multicolumn{1}{c}{Train} & \multicolumn{1}{c}{Val} & \multicolumn{1}{c}{Test} \\ \midrule
\multicolumn{2}{r}{\#Real} & \multicolumn{1}{r}{8,787} & \multicolumn{1}{r}{5,131} & \multicolumn{1}{r}{5,625} & \multicolumn{1}{r}{1,976} & \multicolumn{1}{r}{656} & \multicolumn{1}{r}{661} \\
\multicolumn{2}{r}{\#Fake} & \multicolumn{1}{r}{8,992} & \multicolumn{1}{r}{4,923} & \multicolumn{1}{r}{5,608} & \multicolumn{1}{r}{1,924} & \multicolumn{1}{r}{638} & \multicolumn{1}{r}{628} \\
\multicolumn{2}{r}{Total} & \multicolumn{1}{r}{17,779} & \multicolumn{1}{r}{10,054} & \multicolumn{1}{r}{11,233} & \multicolumn{1}{r}{3,900} & \multicolumn{1}{r}{1,294} & \multicolumn{1}{r}{1,289} \\ \midrule
\multicolumn{2}{r}{\#News Items} & \multicolumn{3}{c}{583,208} & \multicolumn{3}{c}{1,003,646} \\
\multicolumn{2}{r}{\begin{tabular}[c]{@{}r@{}}Min/Avg/Max of\\ $|\mathcal{E}^{mac}|$ in 3 days\end{tabular}} & \multicolumn{3}{c}{41 / 505 / 1,563} & \multicolumn{3}{c}{308 / 1,614 / 2,211} \\
\bottomrule
\end{tabular}
\label{data}%
\vspace{-0.4cm}
\end{table}%

\begin{table*}[t]
    \centering
    \small
    \caption{Performance comparison of base models with and without the NEP. The better result in each group using the same base model are in \textbf{boldface}.}
    \begin{tabular}{clcccccccc}
    \toprule
      \multicolumn{2}{c}{\multirow{2}{*}[-0.3em]{\textbf{Model}}} & \multicolumn{4}{c}{\textbf{Chinese}} & \multicolumn{4}{c}{\textbf{English}} \\
      \cmidrule(lr){3-6} \cmidrule(lr){7-10}
     &   & Acc. & macF1 & F1$_\mathrm{fake}$ & F1$_\mathrm{real}$ & Acc. & macF1 & F1$_\mathrm{fake}$ & F1$_\mathrm{real}$\\
    \midrule
        \multirow{8}{*}{Post-Only} & Bi-LSTM & 0.727 & 0.713 & 0.652 & 0.775 & 0.705 & 0.704 & 0.689 & \textbf{0.719}\\
        &\ \ \  \ \ +\textit{NEP}  & \textbf{0.776} & \textbf{0.771} & \textbf{0.739} & \textbf{0.803} & \textbf{0.718} & \textbf{0.718} & \textbf{0.720} & 0.716\\
        &EANN$_\mathrm{T}$ & 0.732 & 0.718 & 0.657 & 0.780 & 0.700 & 0.699 & 0.683 & 0.714 \\
      & \ \ \  \ \ +\textit{NEP}  & \textbf{0.776} &  \textbf{0.770} &  \textbf{0.733} &  \textbf{0.807} & \textbf{0.722} & \textbf{0.722} & \textbf{0.722} & \textbf{0.722}\\
        &BERT & 0.792 & 0.785 & 0.744 & 0.825 & 0.709 & 0.709 & 0.701 & \textbf{0.716} \\
        &\ \ \  \ \ +\textit{NEP}  & \textbf{0.810} & \textbf{0.805} & \textbf{0.772} & \textbf{0.837} & \textbf{0.718} & \textbf{0.718} & \textbf{0.720} & 0.715\\
        &BERT-Emo & 0.812 & 0.807 & 0.776 & 0.838 & 0.718 & 0.718 & 0.719 & 0.718 \\
        &\ \ \  \ \ +\textit{NEP}  &\textbf{0.831} & \textbf{0.829} & \textbf{0.808} &\textbf{0.850}& \textbf{0.728} & \textbf{0.728} & \textbf{0.728} & \textbf{0.728}\\ \midrule
        \multirow{4}{*}{``Zoom-In''} & DeClarE & 0.764 & 0.758 & 0.720 & 0.795 & 0.714 & 0.714 & 0.709 & \textbf{0.718} \\
        &\ \ \  \ \ +\textit{NEP}  & \textbf{0.800} & \textbf{0.797} & \textbf{0.773} & \textbf{0.822} & \textbf{0.717} & \textbf{0.716}  & \textbf{0.718}  & 0.714\\
        &MAC & 0.755 & 0.751 & 0.717 & 0.784 & 0.706 & 0.705 & 0.708 & 0.701 \\
        &\ \ \  \ \ +\textit{NEP}  & \textbf{0.764} & \textbf{0.760} & \textbf{0.732} & \textbf{0.789} & \textbf{0.716} & \textbf{0.716} & \textbf{0.716} & \textbf{0.716} \\
     \bottomrule
    \end{tabular}
    \label{main_result}
    \vspace{-0.2cm}
\end{table*}


\noindent\textbf{Chinese Dataset}

\textbf{Post:} We merged the non-overlapping parts of multiple Weibo datasets from \cite{majing16} (excluding those unverified), \cite{ced}, \cite{dual-emotion} and \cite{mtm} to achieve a better coverage of years and avoid spurious correlation to specific news environments (e.g., one full of COVID-19 news). To balance the post amount of real/fake classes across the years, we added news posts verified by a news verification system NewsVerify\footnote{\url{https://newsverify.com/}} and resampled the merged set. The final set contains 39,066 verified posts on Weibo ranging from 2010 to 2021.

\textbf{News Environment:} We collected the news items from the official accounts of six representative mainstream news outlets that have over 30M followers on Weibo (see sources in Appendix~\ref{app:sources}). The further post-processing resulted in 583,208 news items from 2010 to 2021.

\noindent\textbf{English Dataset}

\textbf{Post:} Similarly, we merged the datasets from \cite{allinone} (excluding unverified), \cite{multifc} (excluding those without claim dates), and \cite{shaar}. For posts or claims from fact-checking websites, we used the provided claim dates instead of the publication dates of the fact-checking articles, to avoid potential data contamination where the later news environment is more likely to contain corresponding fact-checking news and support direct fact verification. We obtained 6,483 posts from 2014 to 2018 after dropping the posts labeled as neutral and re-sampling.

\textbf{News Environment:} We use news headlines (plus short descriptions if any) from Huffington Post, NPR, and Daily Mail as the substitute of news tweets due to the Twitter's restriction (see sources in Appendix~\ref{app:sources}). The bias rates of the three outlets are respectively left, center, and right according to AllSides Media Bias Chart\footnote{\url{https://www.allsides.com/media-bias/media-bias-ratings}}, for enriching the diversity of news items. We preserved the news headlines from 2014 to 2018 and obtained a set of 1,003,646 news items.

%
%


\subsection{Experimental Setup}
\noindent\textbf{Base Models}
Technically, our NEP could coordinate with any fake news detectors that produce post representation. Here we select four post-only methods and two ``zoom-in'' (knowledge-based) methods as our base models.\footnote{We do not select social context-based methods because it would be impractical to integrate our NEP with them at the cost of timeliness, for the model has to wait for the accumulation of user responses/reposts. We suppose that an asynchronous integration at the \textit{system} level (using post-only/knowledge-based methods with NEP to obtain instant predictions, and update the results later) would be an option, which is beyond our scope.}

\textbf{Post-Only:} 1) Bi-LSTM~\cite{bilstm} which is widely used to encode posts in existing works~\cite{defend, hdsf}; 2)
EANN$_\mathrm{T}$~\cite{eann} which uses adversarial training to remove event-specific features obtained from TextCNN~\cite{textcnn}; 3) BERT~\cite{bert}; 4) BERT-Emo~\cite{dual-emotion} which fuses a series of emotional features with BERT encoded features for classification (publisher emotion version).\footnote{As our work is based on the post text, we use the text-only variant of the original EANN that excludes the image modality and the publisher-emotion-only variant in~\cite{dual-emotion} that excludes the social emotion features.}

\textbf{``Zoom-in'':} 1) DeClarE~\cite{declare} which considers both the post and retrieved documents as possible evidence; 2) MAC~\cite{mac} which build a hierarchical multi-head attention network for evidence-aware detection.

\begin{table*}[t]
    \centering
    \small
    \caption{Performance comparison of the NEP and its variants without the fake news detector or without the environment perception module. The best result in each group is in \textbf{boldface}.}
    \begin{tabular}{lcccccccc}
    \toprule
       \multicolumn{1}{c}{\multirow{2}{*}[-0.3em]{\textbf{Model}}} & \multicolumn{4}{c}{\textbf{Chinese}} & \multicolumn{4}{c}{\textbf{English}} \\
       \cmidrule(lr){2-5} \cmidrule(lr){6-9}
        & Acc. & macF1 & F1$_\mathrm{fake}$ & F1$_\mathrm{real}$ & Acc. & macF1 & F1$_\mathrm{fake}$ & F1$_\mathrm{real}$\\
    \midrule
    	\textsc{MacroEnv}  & 0.689 & 0.659 & 0.557 & 0.761 & 0.693 & 0.693 & \textbf{0.696} & 0.689\\
    	\textsc{MicroEnv}  & 0.666 & 0.626 & 0.503 & 0.748 & 0.695 & 0.695 & 0.694 & 0.696\\
    	\textsc{MacroEnv}+\textsc{MicroEnv}  & \textbf{0.694} & \textbf{0.666} & \textbf{0.569} & \textbf{0.763} & \textbf{0.696} & \textbf{0.696} & 0.694 & \textbf{0.697} \\
    \midrule\midrule
        BERT-Emo + \textit{NEP} & \textbf{0.831} & \textbf{0.829} & \textbf{0.808} &\textbf{0.850}& \textbf{0.728} & \textbf{0.728} & \textbf{0.728} & 0.728\\
        \ \ \ \ \ \textit{w/o} \textsc{MacroEnv}  & 0.822 & 0.819 & 0.794 & 0.843 & 0.726 & 0.726 & 0.726 & 0.725\\
        \ \ \ \ \ \textit{w/o} \textsc{MicroEnv}  & 0.824 & 0.820 & 0.795 & 0.845 & 0.723 & 0.723 & 0.715 & \textbf{0.731}\\
        DeClarE + \textit{NEP} & \textbf{0.797} & \textbf{0.800} & \textbf{0.773} & \textbf{0.822} & \textbf{0.717} & \textbf{0.716}  & 0.718  & \textbf{0.714}\\
        \ \ \ \ \ \textit{w/o} \textsc{MacroEnv}  & 0.776 & 0.771 & 0.735 & 0.806 & 0.712 &	0.711&	0.709&	0.713\\
        \ \ \ \ \ \textit{w/o} \textsc{MicroEnv}  & 0.778 & 0.773 & 0.736 & 0.809 & 0.709&	0.709&	\textbf{0.719}&	0.698\\
        \bottomrule
    \end{tabular}
    \label{ablation}
    \vspace{-0.2cm}
\end{table*}

\noindent\textbf{Implementation Details}
We obtained the sentence representation from SimCSE~\cite{simcse} based on pretrained BERT models in the \textit{Transformers} package~\cite{transformers}\footnote{\textit{bert-base-chinese} and \textit{bert-base-uncased}} and were post-trained on collected news items. We frozed SimCSE when training NEP. 
For DeClarE and MAC, we prepared at most five articles in advance as evidence for each post by retrieving against fact-checking databases.\footnote{We attempted to collect webpages using our posts as queries as~\citet{declare} did but rare ones could serve as evidence except fact-checking articles. As an alternative, we directly used articles from~\cite{mtm} for Chinese and collected \textasciitilde 8k articles from a well-known fact-checking website \url{Snopes.com} for English.}
In environment modeling, $T=3$, $r=0.1$, and $C=22$. We limit $|\mathcal{E}^{mac}|\geq 10$.
We implemented all methods using PyTorch~\cite{pytorch} with AdamW~\cite{adamw} as the optimizer. We reported test results w.r.t. the best validation epoch. Appendix~\ref{app:implementation} provides more implementation details.

\noindent\textbf{Evaluation Metrics.} As the test sets are roughly balanced, we here report accuracy (Acc.), macro F1 score (macF1) and the F1 scores of fake and real class (F1$_\mathrm{fake}$ and F1$_\mathrm{real}$). We will use a new metric for skewed test data (see Section~\ref{online}).

\subsection{Performance Comparison (EQ1)}
\tablename~\ref{main_result} shows the performance of base models with and without the NEP on the two datasets. We have the following observations:

First, with the help of our NEP, all six base models see an performance improvement in terms of accuracy and macro F1.
This validates the effectiveness and compatibility of NEP.

Second, for post-only methods, F1$_\mathrm{fake}$ generally benefits more than F1$_\mathrm{real}$ when using NEP, which indicates that news environments might be more helpful in highlighting the characteristics of fake news. This is a practical property of the NEP as we often focus more on the fake news class.

Third, the ``zoom-in'' knowledge-based methods outperform their corresponding post-only base model (here, Bi-LSTM) with the help of relevant articles, but the improvement is small. This might be led by the difficulty of finding valuable evidence. Our NEP brings additional gains, indicating that the information perceived from news environments is different from verified knowledge, and they play complementary roles.

\subsection{Evaluation on Variants of NEP (EQ2)}


\begin{figure}[t]
\setlength{\abovecaptionskip}{0cm}
\setlength{\belowcaptionskip}{-0.3cm}
\centering
    \subfloat[Proportion Factor $r$]{%
        \includegraphics[width=0.5\linewidth]{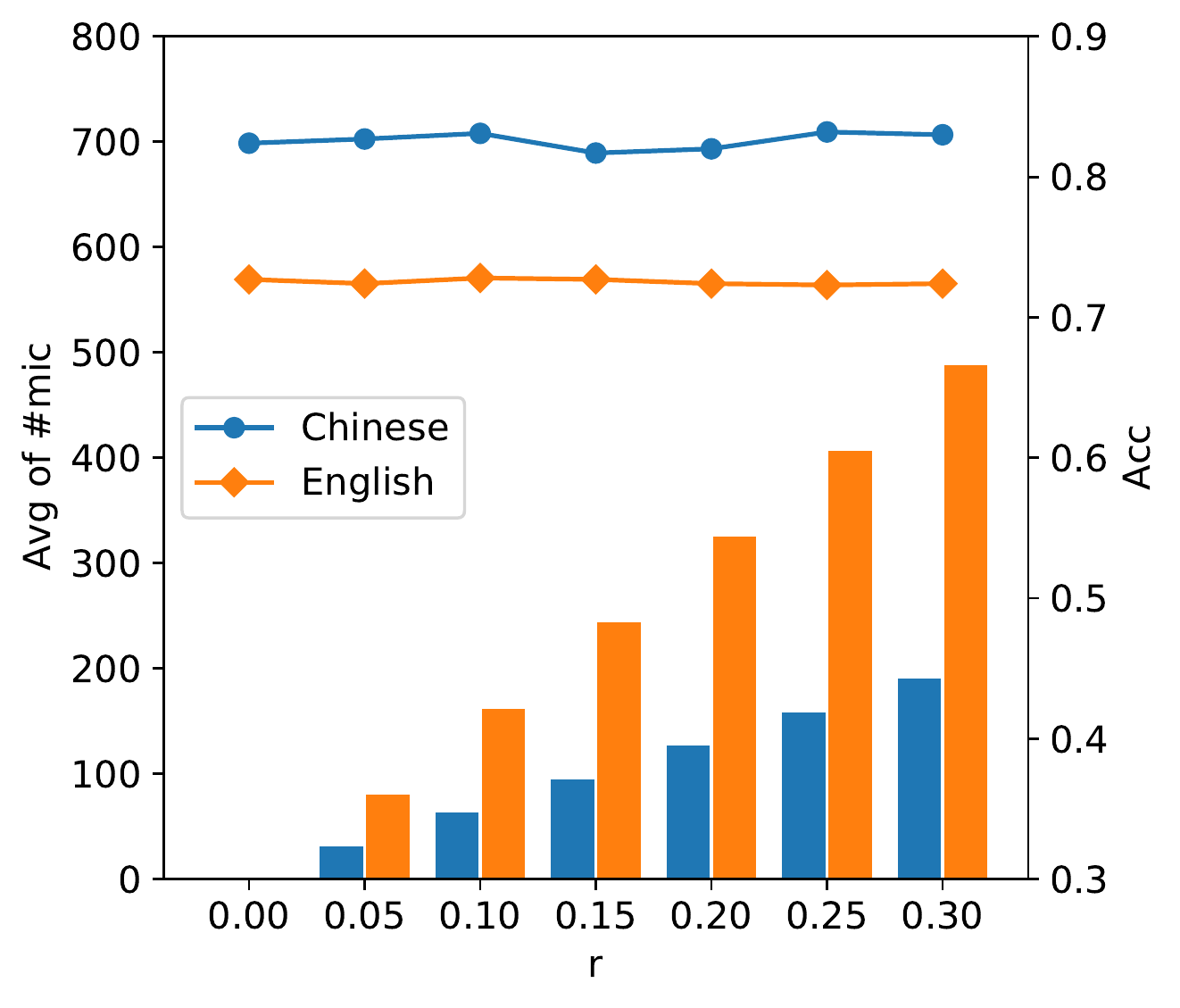}}\hfill
    \subfloat[Day Difference $T$]{%
        \includegraphics[width=0.5\linewidth]{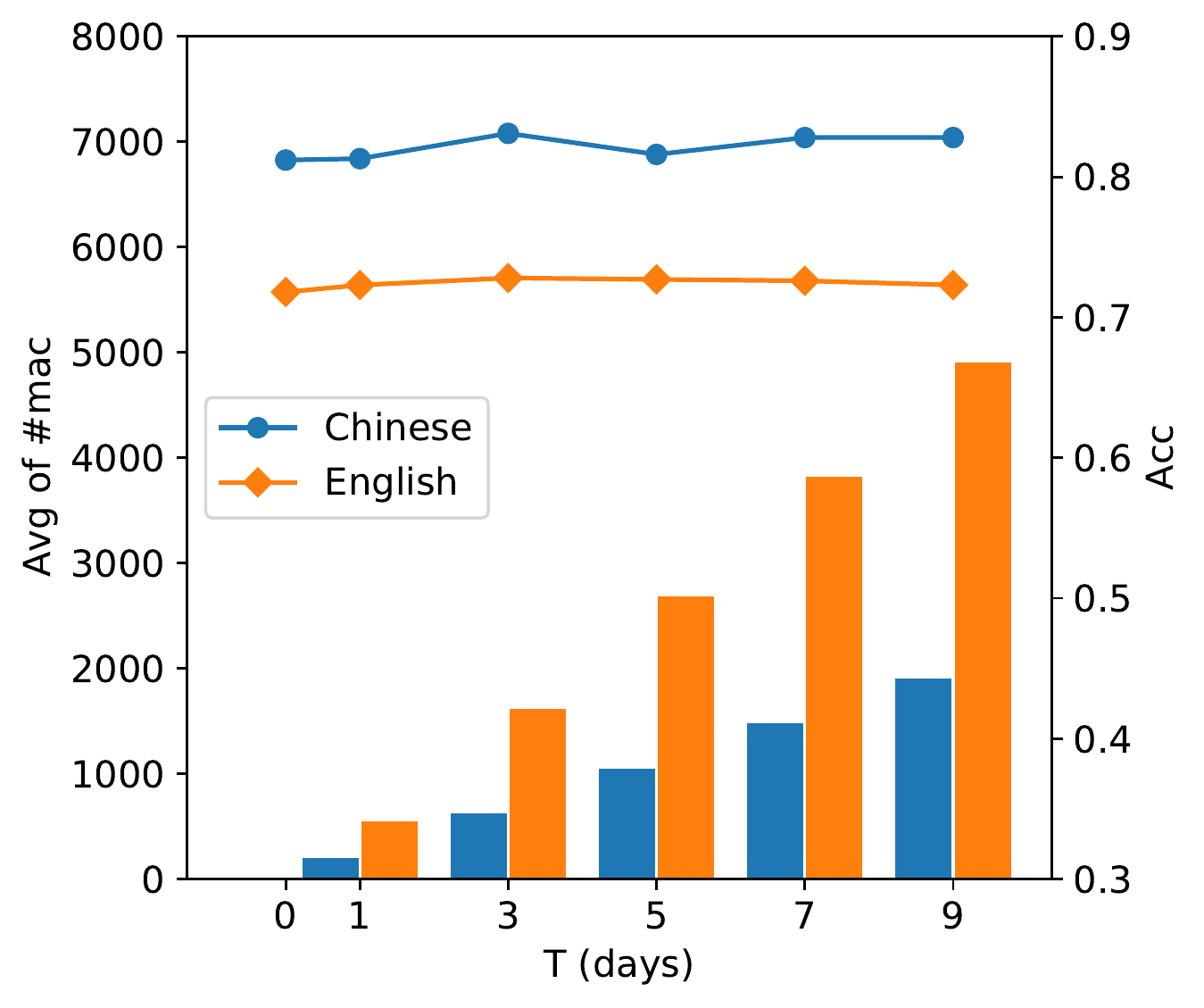}}\hfill
    \caption{Effects of (a) the proportion factor $r$ and (b) the day difference $T$. Lines show the accuracies and bars show the average numbers of news items in the micro/macro environments.}
    \label{fig:rt}
\end{figure}

\noindent\textbf{Ablation Study.} We have two ablative groups as shown in \tablename~\ref{ablation}: 

\textit{w/o Fake News Detector}: We directly use one of the two environment-perceived vectors or both to see whether they can work when not cooperating with the fake news detector's output $\mathbf{o}$. The macro F1 scores on both datasets indicate their moderate effectiveness as sole inputs, and that coordinating with a post-only detector is a more practical setting.

\textit{w/o Environment Perception Modules}: By respectively removing \textsc{MacroEnv} and \textsc{MicroEnv} from the best-performing models BERT-Emo+NEP and DeClarE+NEP, we see a performance drop in macro F1 when removing either of them, indicating that the two environments are both necessary and play complementary roles in detection. 

\noindent\textbf{Effects of the proportion factor $r$ for the \textsc{MicroEnv}.} We adjusted $r$ from 0.05 to 0.30 with a step of 0.05 on BERT-Emo+NEP to see the impact of the scale of the \textsc{MicroEnv} ($T=3$). As \figurename~\ref{fig:rt}(a) shows, the change of $r$ leads to an increase on the size of the \textsc{MicroEnv}, but only fluctuations w.r.t. the accuracy. We do not see significant improvement after $r=0.1$. We speculate that a too small $r$ may hardly cover enough event-similar items while a large $r$ may include much irrelevant information, bringing little gains (e.g., $r=0.3$ in Chinese) or even lowering the performance (e.g., $r=0.15$ for both datasets).

\noindent\textbf{Effects of the day difference $T$ for the \textsc{MacroEnv}.} We set $T=1, 3, 5, 7, 9$ on BERT-Emo+NEP to see how many days of news items to be considered is proper ($T=0$ exactly corresponds to the base model). \figurename~\ref{fig:rt}(b) shows a tendency similar to (a). We find the highest accuracy when $T=3$ on both of the two datasets. This is reasonable as the popularity should be considered in a moderately short time interval to allow the events to develop but not to be forgotten.

\begin{figure}[t]
\setlength{\belowcaptionskip}{-0.5cm}
\centering
    \includegraphics[width=.9\linewidth]{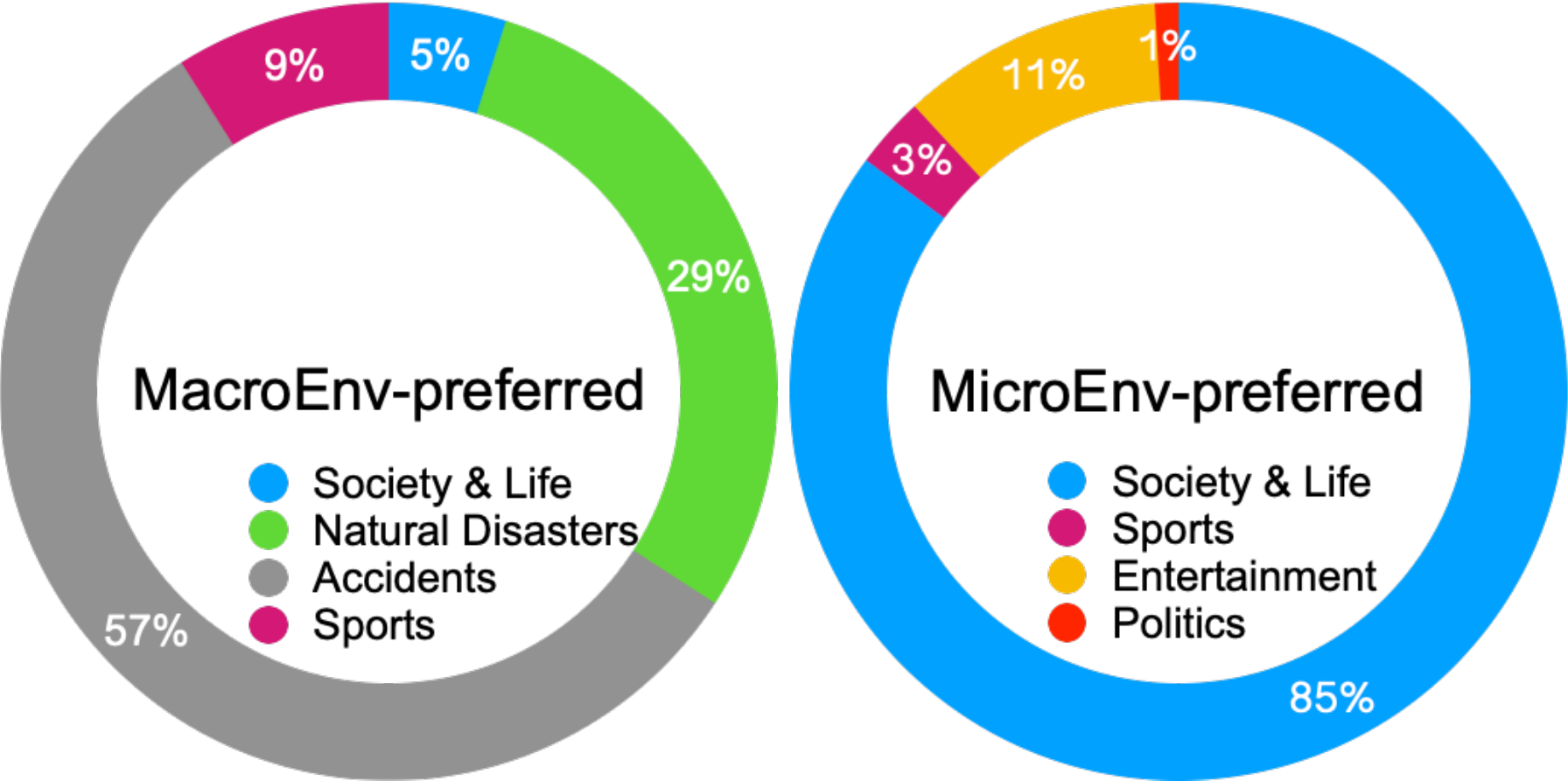}
    \caption{Categories of \textsc{MacroEnv}- and \textsc{MicroEnv}-preferred samples.}
    \label{fig:macro_vs_micro}
\end{figure}

\begin{figure*}[t]
\setlength{\abovecaptionskip}{0cm}
\setlength{\belowcaptionskip}{-0.3cm}
\centering
    \includegraphics[width=\linewidth]{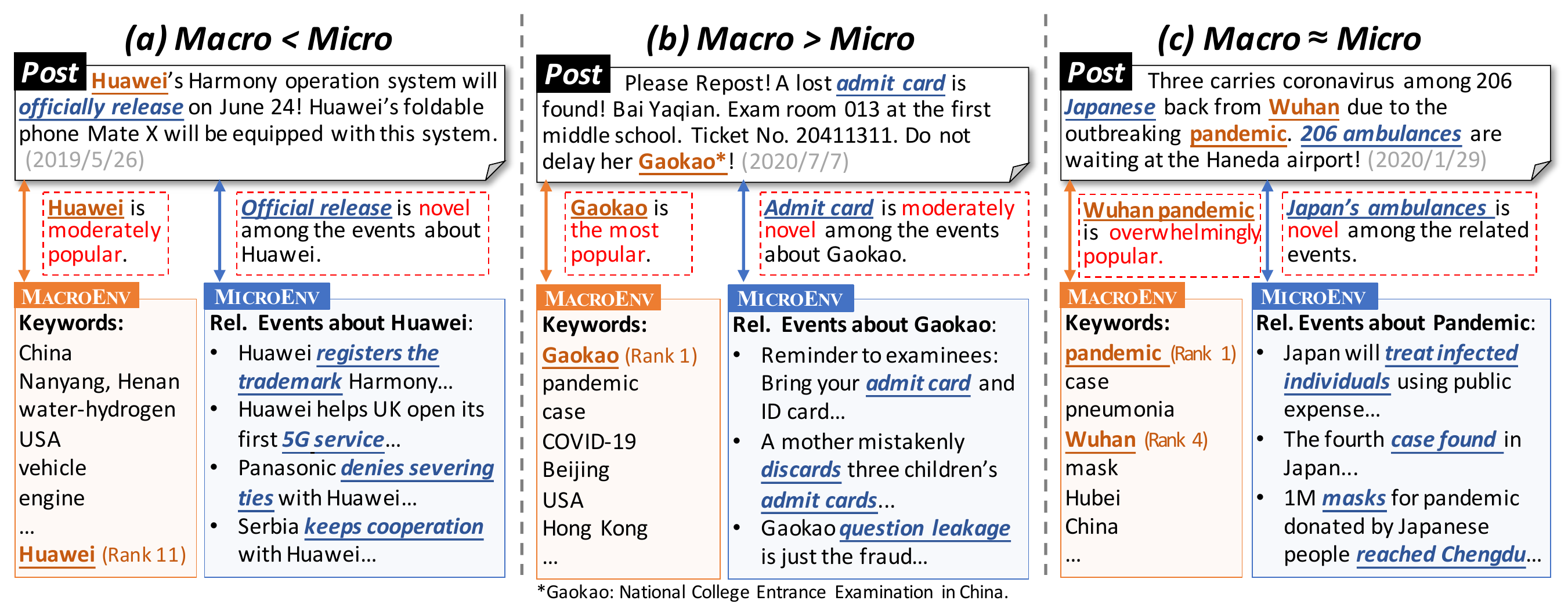}
    \caption{Three fake news cases with different preferences on environmental information. {\color{E66826} \underline{Underlined regular}} words hit the keywords in the \textsc{MacroEnv} and {\color{345AB5} \underline{\textit{underlined italic}}} words are related to the \textsc{MicroEnv}. Keywords are extracted using TextRank~\cite{textrank}.}
    \label{fig:cases}
\end{figure*}

\subsection{Environment Analysis (EQ3)}



\noindent\textbf{Categorization of macro- and micro-preferred samples.} We selected the top 1\% of Chinese fake news samples which NEP relies more on \textsc{MacroEnv} or \textsc{MicroEnv} according to the gate vectors. Then we manually categorized these samples to probe what information the macro/micro environment might provide. From \figurename~\ref{fig:macro_vs_micro}, we see that \textsc{MacroEnv} is more useful for samples about natural disasters and accidents (e.g., earthquakes and air crashes), while \textsc{MicroEnv} works effectively in Society \& Life (e.g., robbery and education). This is in line with our intuition: \textsc{MacroEnv}-preferred fake news posts are often related to sensational events, so the popularity in \textsc{MacroEnv} would help more; and \textsc{MicroEnv}-preferred ones are often related to common events in daily news, and thus its novelty in \textsc{MicroEnv} would be highlighted. This analysis would deepen our understanding on the applicability of different news environments.

\noindent\textbf{Case study.}
\figurename~\ref{fig:cases} shows three fake news cases in different scenarios. Case (a)  relies more on \textsc{MicroEnv} than \textsc{MacroEnv}. We can see moderate popularity of its event about Huawei but the message about HarmonyOS is novel among the items on the 5G and cooperations. In contrast, the admit card in case (b) is moderately novel but Gaokao is the most popular event, so the NEP puts higher weight on \textsc{MacroEnv}. Case (c) is a popular and novel fake news about Japan's great healthcare for citizens coming back from Wuhan which is posted during the first round of COVID-19 pandemic in China. The exploitation of both-side information makes a tie between the two environments. These cases intuitively show how NEP handles different scenarios. We incorporate further analysis on the case that the news environment might be ineffective in Appendix~\ref{app:failure}.

\begin{figure}[t]
\setlength{\abovecaptionskip}{0cm}
\setlength{\belowcaptionskip}{-0.4cm}
\centering
    \includegraphics[width=0.75\linewidth]{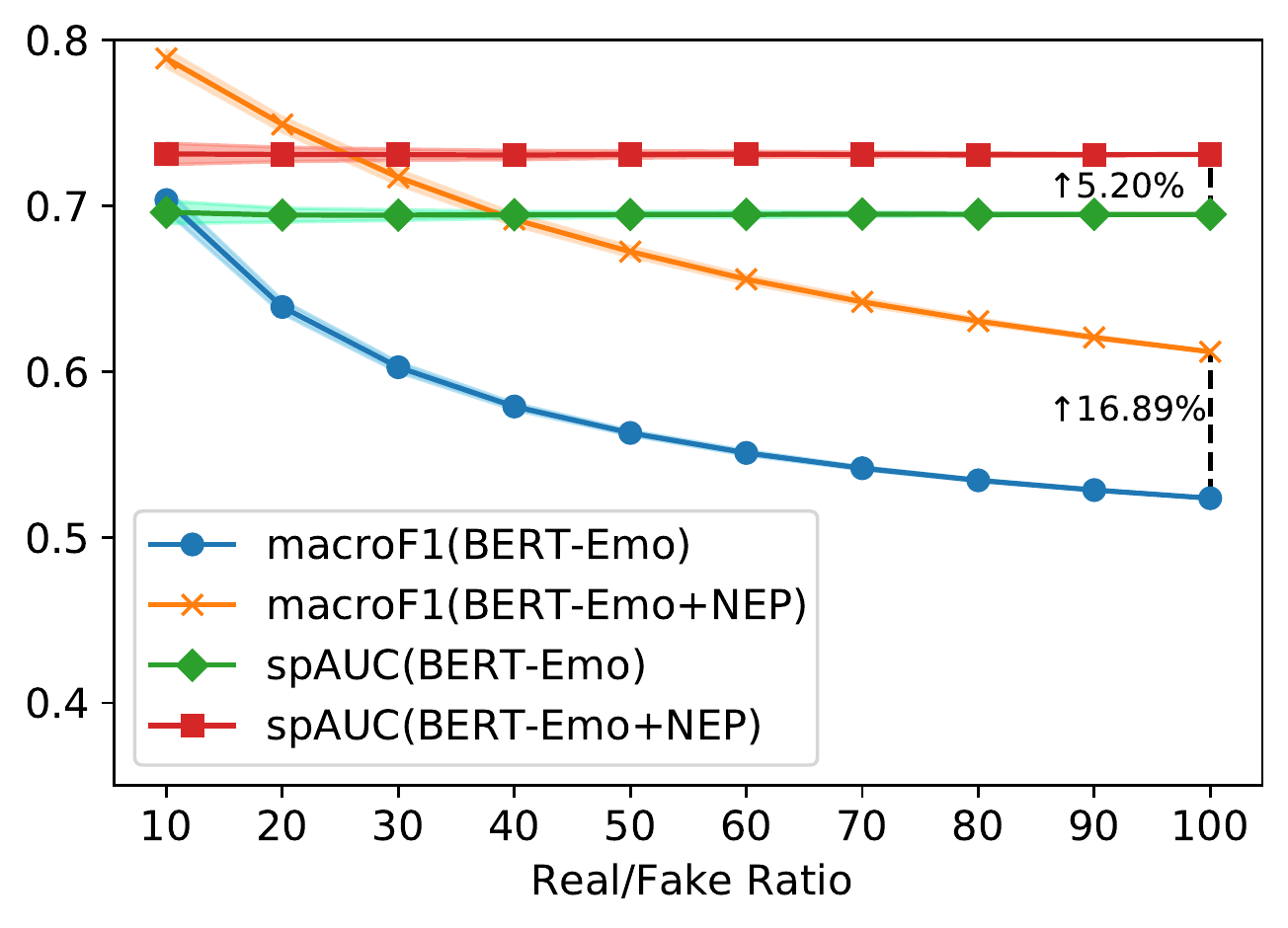}
    \caption{Macro F1s and spAUCs on the online data in different real/fake ratios. We sampled 100 times from the 100:1 set for each fo the first nine ratios. Shadows show the standard deviations. The percentages denote relative improvements using the NEP.}
    \label{fig:ol}
\end{figure}

\section{Discussion in Practical Systems}
\label{online}
\noindent\textbf{Evaluation on skewed online data.}
We tested BERT-Emo and BERT-Emo+NEP on a dump of seven-month data from a Chinese fake news detection system. Different from offline datasets, this real-world set is highly skewed (30,977 real vs. 309 fake, roughly 100:1).\footnote{The online test set and the offline sets do not intersect.} Under such skewed circumstance, some metrics we used in Tables~\ref{main_result} and~\ref{ablation} could hardly show the differences of performances among models (e.g., a model predicting all samples as real will have an incredible accuracy of 0.990). Here, we report macro F1 and standardized partial AUC with false positive rate of at most 0.1 (spAUC$_{\mathrm{FPR} \leq 0.1}$, \citealp{spauc}, see Appendix~\ref{app:spauc} for the calculation detail) under different real/fake ratios (from 10:1 to 100:1). As shown in \figurename~\ref{fig:ol}, NEP brings relative improvements of 16.89\% and 5.20\% in macF1 and spAUC$_{\mathrm{FPR} \leq 0.1}$, showing its effectiveness in skewed, real scenarios.

\noindent\textbf{Friendliness to Practical Systems.}
The NEP is not only a new direction for fake news detection but also inherently friendly to practical systems: \textbf{1) Timeliness.} Our NEP works \emph{instantly} as it only requires the post and mainstream news published a few days before. In practice, a system would not construct the required collection on demand but prepare it ahead by maintaining a queue of news items. \textbf{2) Compatibility.} Our perception module can be integrated with existing methods, which we validated on six representative ones (\tablename~\ref{main_result}). \textbf{3) Data Accessibility.} The data to construct news environments is easy to access, especially compared with obtaining credible knowledge sources. The advantages may encourage the deployment of NEP into practical systems.

\section{Conclusion and Future Work}
We proposed the NEP to observe news environments for fake news detection on social media. We designed popularity- and novelty-oriented perception modules to assist fake news detectors. Experiments on offline and online data show the effectiveness of NEP in boosting the performance of existing models. We drew insights on how NEP help to interpret the contribution of macro and micro environment in fake news detection.

As this is the first work on the role of news environments for fake news detection, we believe further exploration is required for a deeper understanding of the effects of news environments and beyond. In the future, we plan to explore: 1) including historical news or background to handle posts weakly related to the present environment; 2) modeling post-environment relationships with diverse similarity metrics or even from other perspectives; 3) investigating the effects of different news environments (e.g., biased vs. neutral ones) to make the environment construction more principled; 4) extending this type of methodology from the text-only detection to multi-modal and social graph-based detection.

\section*{Acknowledgements}
The authors thank Guang Yang, Peng Qi, Zihao He, and anonymous reviewers for their insightful comments. This work was supported by the Zhejiang Provincial Key Research and Development Program of China (No. 2021C01164).

\section*{Ethical Considerations}

\noindent\textbf{Application.} Our framework does not present direct societal consequence and is expected to benefit the defense against the fake news issue. It can serve as a detection module for fake news detection systems, especially when the given post is closely related to the events that happened recently, with no need to wait for the accumulation of user responses or query to knowledge sources. Due to the requirement of real-time access to open news sources (source list can be determined as needed), it might be easier to deploy for service providers (e.g., news platforms) and media outlets.

\noindent\textbf{Data.} Our data is mostly based on existing datasets, except the news items for constructing news environments. All news items (or headlines) are open and accessible to readers and have no issues with user privacy. The media outlets in the English dataset might be considered ``biased'', so we carefully select a left, a center, and a right outlet (whose headlines are available) according to the AllSides Media Bias Chart. In China, a media outlet might be state-run (e.g., CCTV News), local-government-run (e.g., The Paper), or business-run (e.g., Toutiao News). With no widely recognized bias chart of Chinese media as a reference, we select media outlets based on their influence (e.g., number of followers) on Weibo from the three categories for the sake of representativeness.

\bibliography{custom.bib}
\bibliographystyle{acl_natbib}


\appendix
\balance

\section{Sources of News Items as the Environmental Elements}
\label{app:sources}
\tablename~\ref{tab:sources} shows the selected news outlets that provides news items as the elements for news environment construction in Chinese and English. The Huffington Post part was derived from the Kaggle page~\citep{misra2018news,misra2021sculpting} and we crawled the other parts.

\begin{table*}[t]
\caption{Sources of News Items in the Chinese and English datasets.}
\label{tab:sources}
{\small
\begin{tabular}{p{0.18\linewidth}p{0.73\linewidth}}
\toprule
\textbf{News Outlet}  & \textbf{URL} \\
\midrule
\multicolumn{2}{l}{\textbf{\textit{Chinese}}} \\ 
People's Daily & \url{https://weibo.com/u/2803301701} \\
Xinhua Agency & \url{https://weibo.com/u/1699432410} \\
Xinhua Net & \url{https://weibo.com/u/2810373291} \\
CCTV News & \url{https://weibo.com/u/2656274875} \\
The Paper & \url{https://weibo.com/u/5044281310} \\
Toutiao News & \url{https://weibo.com/u/1618051664} \\
\midrule
\multicolumn{2}{l}{\textbf{\textit{English}}} \\ 
Huffington Post & \url{https://www.kaggle.com/rmisra/news-category-dataset/} \\
NPR & \url{https://www.npr.org/sections/news/archive} \\
Daily Mail & \url{https://www.dailymail.co.uk/home/sitemaparchive/} \\
\bottomrule
\end{tabular}
}
\end{table*}

\section{Supplementary Implementation Details}
\label{app:implementation}
\subsection{Kernel Settings}
\label{app:kernel}
We use $C=22$ kernels for softly counting the cosine similarities. Following~\cite{kernelpooling}, we first determine 21 kernels whose $\mu$s scatter in $[-1,1]$ with an interval of 0.1 and $\sigma^2$s are all 0.05. Then we add a kernel with a $\mu$ of 0.99 and a $\sigma^2$ of 0.01, specially for extremely similar situations. The final kernel list is [(-1.0, 0.1), (-0.9, 0.1),$\cdots$, (1.0,0.1), (0.99, 0.01)]

\subsection{Post-Training SimCSE}
We post-trained the BERT models for two epochs, with the temperature coefficient $\tau$ of 0.05, the dropout rate of 0.3 (Chinese, hereafter, C) and 0.1 (English, hereafter, E), and the maximum length of 256 (C) and 128 (E).

\subsection{Implementation of Base Models}
\begin{compactitem}
	\item \textbf{Bi-LSTM}: The hidden dims are 128 (C) and 256 (E). The maximum lengths are 256 (C) and 128 (E). The number of layers are 1 (C) and 2 (E). We use \textit{sgns.weibo.bigram-char}\footnote{\url{https://github.com/Embedding/Chinese-Word-Vectors}}~\cite{chinese-embedding} for Chinese and \textit{glove.840B.300d}\footnote{\url{https://nlp.stanford.edu/projects/glove/}}~\cite{glove} for English to obtain the word embeddings. The Chinese texts are segmented using \textit{jieba}\footnote{\url{https://github.com/fxsjy/jieba}} and the English texts are tokenized using \textit{NLTK}~\cite{nltk}.
	\item \textbf{EANN$_\mathrm{T}$}: The hidden dims, maximum lengths, and word embeddings are the same as Bi-LSTM. The kernel size for both datasets are $[1,2,3,4]$. The numbers of filters are 20 (C) and 30 (E). We ran K-means~\cite{kmeans} in the \textit{scikit-learn} package to gather the training samples into 300 clusters (corresponding to 300 events).\footnote{\url{https://scikit-learn.org/stable/modules/generated/sklearn.cluster.KMeans.html}}
	\item \textbf{BERT} and \textbf{BERT-Emo}: We use \textit{bert-base-chinese} and \textit{bert-based-uncased} for Chinese and English, respectively. The maximum lengths are 256 (C) and 128 (E). The dimension of each token representation is 768. 
	\item \textbf{DeClarE} and \textbf{MAC}: The Bi-LSTM component keeps the same settings as the post-only Bi-LSTM. The maximum lengths of articles are 100 (C) and 256 (E).

\end{compactitem}

\section{Calculation of spAUC}
\label{app:spauc}
Real-world fake news detection systems inevitably face a challenge of high imbalance of data (\#real$>>$\#fake), even if pre-screening procedures like check-worthiness estimation~\cite{claimbuster, checkthat18} are equipped. In the online test, we use the standardized partial AUC (spAUC)~\cite{spauc} for evaluation. It is suitable to our scenario where we expect the method to find fake news posts as many as possible with an acceptable misclassification rate of real ones. The partial AUC over the false positive rate $[0,x]$ is:
\begin{equation}
	\mathrm{pAUC}_{\mathrm{FPR}\leq x} = \int_{0}^{x}\mathrm{ROC}(x)\ \mathrm{d} x,
\end{equation}
where $\mathrm{ROC}$ is the Receiver Operating Characteristic curve.  The spAUC is calculated as
\begin{equation}
	\mathrm{spAUC}_{\mathrm{FPR}\leq x}=\frac{1}{2}\left(1+\frac{\mathrm{pAUC}_{\mathrm{FPR}\leq x}-\frac{1}{2}x^2}{x-\frac{1}{2}x^2}\right).
\end{equation}
In our experiment, we use the implementation in the \textit{scikit-learn} package.\footnote{\url{https://scikit-learn.org/stable/modules/generated/sklearn.metrics.roc_auc_score.html}}

\section{Analysis on the Case Weakly Related to News Environments}
\label{app:failure}

\begin{figure}[htbp]
\centering
    \includegraphics[width=0.85\linewidth]{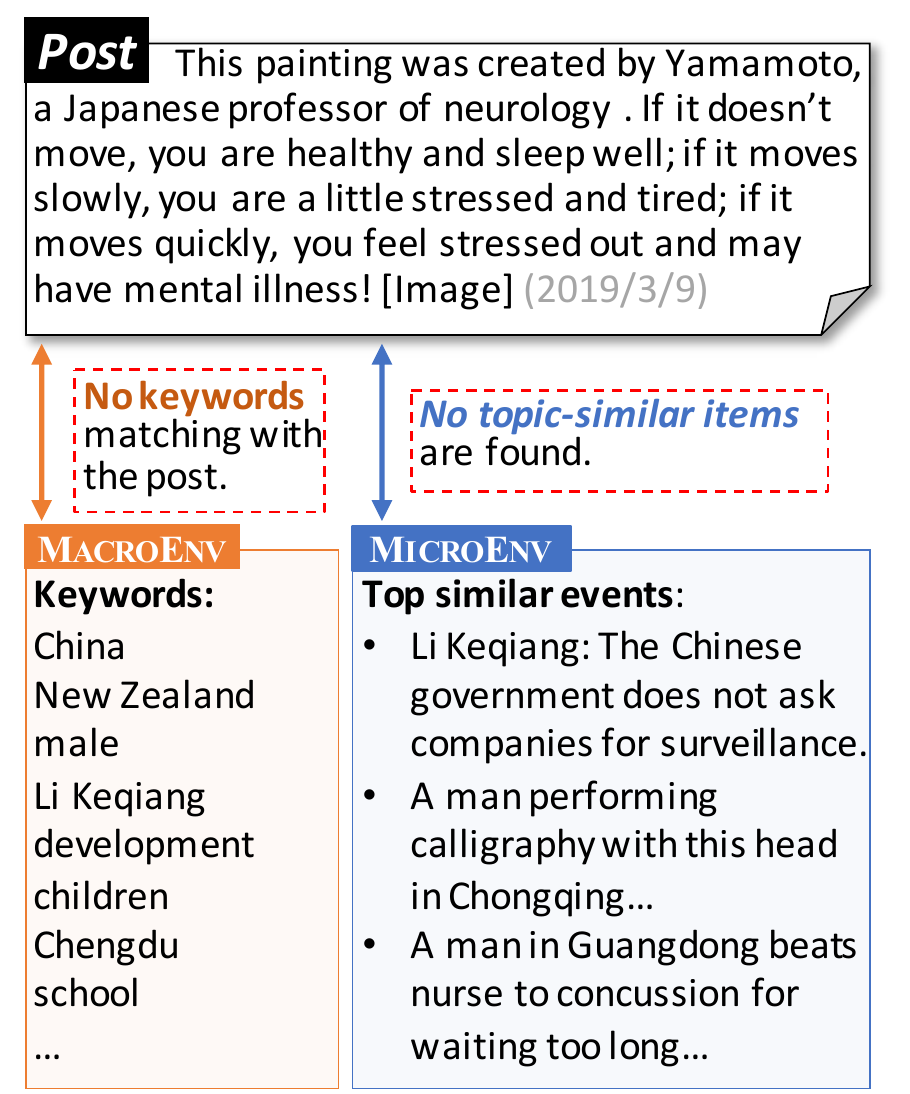}
    \caption{A case that is weakly related to its news environment.}
    \label{fig:app_case}
\end{figure}

\figurename~\ref{fig:app_case} shows a case that is weakly related to its news environment. Its words have no intersection with the keywords in the macro environment and the top similar events seem not very related. In this case, our NEP has limited utility as its nature of recency. That might explain why the performances were mostly lower than the post-only methods when we evaluated the NEP alone.
For this case, it actually has some novelty (a novel and simple test of personal stress)  but is involved with a long-lasting discussed topic---mental health, instead of a hot event being discussed at the very moment.
This inspires us to explore how to incorporate more historical and background references to build a comprehensive understanding of the connection between a fake news post and broader societal environments in the future.

\end{document}